\documentclass[sigconf]{acmart}
\renewcommand\footnotetextcopyrightpermission[1]{}
\usepackage{pgfplots}
\pgfplotsset{width=10cm,compat=1.9}

\newcommand \Workers {W}
\newcommand \worker {w}
\newcommand \Tasks {S}
\newcommand \task {s}

\newcommand \Skills {M}
\newcommand \skill {\mu}
\newcommand \Reqs {\Lambda}
\newcommand \req {\lambda}

\begin{document}






%

\title{Learning to match}

%
%
%
%
%

\author{Philip Ekman}
\affiliation{Chalmers University of Technology}
\email{ekman1991@gmail.com}
\author{Sebastian Bellevik}
\affiliation{Chalmers University of Technology}
\email{bellevik@gmail.com}
\author{Christos Dimitrakakis}
\affiliation{University of Lille\\Harvard University\\Chalmers University of Technology}
  \email{christos.dimitrakakis@gmail.com}
\author{Aristide Tossou}
\affiliation{Chalmers University of Technology}
\email{aristide@chalmers.se}


\maketitle
\begin{abstract}
Outsourcing tasks to previously unknown parties is becoming more common. One specific such problem involves matching a set of workers to a set of tasks. Even if the latter have precise requirements, the quality of individual workers is usually unknown. The problem is thus a version of matching under uncertainty. We believe that this type of problem is going to be increasingly important.

When the problem involves only a single skill or type of job, it is essentially a type of bandit problem, and can be solved with standard algorithms. However, we develop an algorithm that can perform matching for workers with multiple skills hired for multiple jobs with multiple requirements. We perform an experimental evaluation in both single-task and multi-task problems, comparing with the bounded $\epsilon$-first algorithm, as well as an oracle that knows the true skills of workers. One of the algorithms we developed gives results approaching 85\% of oracle's performance. We invite the community to take a closer look at this problem and develop real-world benchmarks.
\end{abstract}


\section{Introduction}
A trend that has been observed over the recent years is that many companies, especially software companies, are outsourcing work to previously unknown parties \cite{tran2014efficient}. Instead of outsourcing tasks to known, or previously used, firms, they obtain workers through online platforms such as \textit{Amazon's Mechanical Turk} \cite{mechanical_turk}. This type of outsourcing is sometimes referred to as crowdsourcing. On the market today there are several crowdsourcing platforms that outsource simpler tasks which can be completed during a short period of time and without a specific set of high level skills \cite{many_crowd_out_sourcing}.

\paragraph{Previous work} There are many methods for the optimal matching of workers and tasks when all variables are known~\cite{multiple_algorithms}, including for online matching problems in the case of bipartite graphs~\cite{online_matching}. The main novelty in our setting is the existence of \emph{unknown} variables. These create  additional uncertainty, which significantly complicates the problem. The second aspect in which we differ from bipartite matching is that each task may require a specific set of skills, each with a certain amount of experience, to be completed.

\paragraph{Contributions} Many matching platforms allow workers to
enter their own perceived skill levels as an input. However, the
workers may not be truthful. While this can be avoided by only assigning simple tasks that almost anyone can do, it does not allow for the completion of complex tasks. In this paper we consider the problem of optimal task allocation to workers with unknown skills, so as to maximise the number of tasks solved over time. This is achieved by introducing some algorithms for the problem of matching under uncertainty. 

\subsection{Problem description}

The problem consists of a number of workers and tasks (or jobs) to be
matched in an ideal way. Every task has a number of skill level
requirements, representing what is needed from a worker to complete
it. We model this through a set of workers $\Workers$ and their
associated skills $\Skills$, as well as a set of tasks $\Tasks$. Every
worker $\worker \in \Workers$ has a set of skills $\skill_\worker$,
each with an unknown true skill level $\skill_{\worker,m}$. We denote the
complete description of skills levels of all workers simply by
$\skill$.

Employers, on the other hand, set precise skill level requirements for
each task, which we are known to the matching algorithm.  More
precisely, each task $\task \in \Tasks$ has a set of requirements
$\Reqs_\task$ over skills, each with a requirement level
$\req_{\task,m} \in [0,1]$ for each skill $m$. After a set of workers
is assigned to a set of tasks, the employers report on the workers'
performance, which depends on the difference between $\skill$ and
$\req$. This feedback is then used by the matching algorithm to assign
workers in the next cycle.

The algorithm's action $a_k \in A$ in iteration $k$ is an assignment
that matches workers with tasks. This results in a nonnegative reward,
$r_k \geq 0$ with distribution $r_k \sim P(a, \skill, \Tasks)$, with
expectation $E(r | a, \skill, \Tasks)$, that depends on the assignment and the
set of workers and tasks available. More precisely, each action $a_k$
is a set of assignments $(\worker,\task)$, and the reward received is the sum of
rewards obtained for each worker
\begin{equation}
r_k = \sum_{(\worker,\task) \in a_k} r_{k,\worker}, \qquad r_{k,\worker} \sim P(r \mid \skill_{\worker}, \task),
\end{equation}
i.e. the reward obtained for worker $\worker$ depends on her skill level $\skill_\worker$ and the tasks $\task$ she is assigned to.  This reward is calculated by comparing the skill levels of the workers with the requirement levels of the task, where a skill level higher than the corresponding requirement level means there is a higher probability of success. 
Hence, we can state the goal as maximising the total reward gained from assigning tasks to workers,
$\sum_k r_k$.

When the actual worker skills are known, it is possible to attain the
optimal solution through e.g. the Hungarian
Algorithm~\cite{hung_steps}. However, as the characteristics of the
workers are unknown we must both estimate their true skill levels, as
well as try to match them to jobs as well as possible given our
uncertainty.  In the simplest case, we can use point estimates
$\hat{\skill}_{\worker,m}$ for each worker's skills. In general, we
shall use $\widehat{\skill}$ and $\widehat{\skill_{\worker,m}}$ to
denote our estimate of worker skills.  The estimated skill levels will
be gradually updated by the algorithm and ideally we would like
$\hat{\skill}$ to converge to $\skill$. However, it is not necessary
to learn the true skills for all workers, just enough to always pick
the best workers for the tasks at hand. The actual estimation depends
on the feedback model we have. In this paper, we limit ourselves to
the following simple feedback.

\paragraph{Feedback model}
After each completed task, the information about the worker known by the algorithm is updated. For each task our assigned workers perform, we obtain a reward $r_{\worker,\task}$. This reward reflects whether the worker was able to perform the task. In our model, the reward only depends on whether the worker's skill is sufficient for the task.
\begin{equation}
\label{eq:skill_est}
r_{\worker,\task} = \quad \begin{cases}
      \textrm{Bernoulli}(1-p), & \skill_{\worker,\task} \geq \req_{\task} \\
      \textrm{Bernoulli}(p), & \textrm{otherwise}
    \end{cases},
\end{equation}
where $\req_{\worker}$ is the task's requirement level for requirement
$i$.  In this model, workers either fail or succeed in a task, and
better workers have a higher probability of success.

\paragraph{Assignment model}
We could consider either a bipartite matching or an unrestricted
matching.  In unrestricted matching, a single worker can be assigned
to multiple tasks. When the tasks provided are simple and take no more
than a couple of seconds of a worker's time\footnote{c.f. the Get
  Another Label project} this might be realistic.  However, in many
real world scenarios tasks will take longer to complete. In the second
scenario we use bipartite matching, the number of workers are the same
as the number of tasks. We believe that both assignment models are
important, and hence consider both of them in this paper.

\paragraph{Practical considerations and interpretation}
If the number of available tasks is much greater than that of workers, then it is easier to estimate each worker's skill by repeated assignment to tasks. However, this is not always the case in a real world setting, where the number of available workers can be much higher than the number of available tasks. To circumvent this problem, companies can introduce \emph{virtual tasks}. In a real setting, these can be anything from unimportant tasks that companies are willing to waste, to simple tests, all with the purpose of estimating worker skill levels. While in this paper we make no distinction is made between \emph{virtual} and \emph{real} tasks, we model their existence by assuming that there are more available tasks (in total) than workers. Finally, to move from the unrestricted to the bipartite setting, we can simply split a large number of tasks into blocks equalling the size of the worker pool.

\section{Related work}
\label{sec:related-work}
Even though this specific setting has not been studied before,
algorithms applicable to similar problems could also be useful
here. One of the simplest one is $\epsilon$-greedy selection
(Section~\ref{sec:epsilon-greedy}). In our setting, this either
assigns the \emph{apparently} best workers to tasks, or (with
probability $\epsilon$) performs a random assignment. However, unless
$\epsilon$ is appropriately tuned, its behaviour is far from
optimal. A well-known algorithm for bandits, UCB
(Section~\ref{sec:ucb}), is directly applicable in this setting,
whenever we only have one skill to consider. A similar algorithm,
bounded $\epsilon$-first, is applicable in the case where we have a
budget for workers. The Hungarian algorithm, described in Section
\ref{sec:hungarian}, is an efficient way to perform the optimal
matching whenever we have perfect knowledge of worker skills. However,
since we do not actually know the parameters, this problem can be seen
as similar to that of contextual combinatorial bandits, described in
Section~\ref{sec:contextual-combinatorial-bandits}.

\subsection{Epsilon greedy}
\label{sec:epsilon-greedy}
This algorithm~\cite{epsilon-greedy} choose an apparently best action most of the time, and a random action with probability $\varepsilon$. In our setting, this means that with probability $\varepsilon$ we choose $a \in A$ uniformly. Otherwise, given our estimate $\widehat{\skill}$ of skill levels, we choose the action maximising expected reward assuming our estimate is correct, i.e.
\begin{equation}
a_k \in \arg\max_{a \in A} E(r \mid a, \widehat{\skill}, \Tasks)
\end{equation}
Finally, instead of a fixed amount of randomness, we can select $\varepsilon_{k} = d^k \varepsilon_{k-1}$, where $\varepsilon_{k}$ is the current epsilon, $d_{k}$ is the drop rate to the power of the number of tasks completed and $\varepsilon_{k-1}$ is the epsilon during the previously performed task. As more tasks are performed, the probability to pick a random worker should decrease, because the algorithm should have learned something about the available workers.

\subsection{Upper confidence bound (UCB)}
\label{sec:ucb}
UCB~\cite{ucb} is an algorithm for near-optimal exploration in bandit problems, and avoids the problem of  selecting a rate for decreasing $\varepsilon_k$. When we have a single worker to select, it is possible to select workers by looking at their average performance plus some confidence bound expressing our uncertainty about them. We can do this by selecting an assignment maximising
\begin{equation}
\sum_{\worker,\task} \widehat{r_{\worker,\task}} + \sqrt{\dfrac{2ln(n)}{n_{\worker}}},
\end{equation}
Where $j$ is the current worker, $\widehat{r_{\worker,\task}}$ the average reward for worker $\worker$, $n$ is the total number of tasks performed and $n_{j}$ is the number of tasks performed by worker $j$. Although this algorithm is directly applicable to the the case when we only have one type of task, we can also apply it to the multiple task case by simply looking at the average reward obtained over all tasks. This, however, ignores a lot of information and will result in suboptimal performance.

\subsection{Bounded epsilon first}
\label{sec:match_bef}
 Bounded Epsilon First (BEF)~\cite{tran2014efficient} is a bandit algorithm for a fixed budget, where selecting a worker incurs a cost. The algorithm consists of an exploration phase and an exploitation phase. A certain part, $\varepsilon \in [0,1]$, of the budget, $B$, is dedicated to the exploration phase. During that phase, each worker is assigned a task and is paid, until $\varepsilon$B is depleted. This phase is used to estimate the workers skill levels. The skill levels are later used to determine which worker to use for a certain task during the exploitation phase.

The exploitation phase is used to select the best workers available for each task. Because the algorithm is unaware of the true skill levels for the workers, it uses the estimated skill levels obtained through the exploration phase. During this phase the bounded knapsack algorithm~\cite{bounded_knapsack_problem} is used to select the best available worker for the task. First, the workers are sorted by their density $\delta_{\worker} = \dfrac{\widehat{\skill_{\worker}}}{c_{\worker}}$,
where $\widehat{\skill_{\worker}}$ is the estimated skill level for worker $\worker$ and $c_{\worker}$ is their cost. After sorting the list of workers, the algorithm pulls the arms of the worker with the highest density until the limit for that worker is reached. This is repeated until the rest of the budget, $1-\varepsilon$, is depleted.

\subsection{Hungarian algorithm}
\label{sec:hungarian}
The Hungarian Algorithm \cite{hung_alg} is a polynomial time algorithm for bipartite matching problems. It uses a cost matrix, $C$, of size $n \times n$, where $n$ is the number of workers and tasks, and each entry in the matrix is nonnegative. The entry $C_{\worker,\task}$  represents the cost of assigning worker $\worker$ to task $\task$. Given this matrix the algorithm finds the optimal way of assigning workers to each task. In our case, we can set $C_{\worker,\task}$ to be equal to the negative expected reward of assigning a worker to a task. If we use the \emph{actual} skill levels, then this corresponds to an oracle algorithm. If we use the \emph{estimated} skill levels, then we obtain a simple greedy assignment algorithm.

\subsection{Contextual combinatorial bandits}
\label{sec:contextual-combinatorial-bandits}
In the contextual combinatorial bandit problem\cite{qin2014contextual}, at time $t$ the decision maker observes a contextual vector $\boldsymbol{\req}_t$, and has a choice of $K$ arms to pull. He selects a subset $S_t \subset 2^{[K]}$ and, for each arm $j \in S_t$, obtains a reward $r_{t,i}$ with expectation $\boldsymbol{\skill}^\top \boldsymbol{\req}_t(j)$, where $\boldsymbol{\skill}$ is an unknown parameter vector. While this setting is quite close to ours, note that the main difference is that the decision maker must match jobs to workers, rather than simply select workers for the same task. So, this would correspond to a selection $(i,j)$ being made, with $j$ being the job and $i$ the worker, and where the reward for each job $j$ being $\boldsymbol{\skill}^\top(i) \boldsymbol{\req}_t(j)$. 
Thus, contextual bandit algorithms are not immediately extensible to our setting.

\section{Algorithms}

Existing algorithms cannot be applied directly to this problem: they are either designed for a restricted problem setting, like multi-armed bandits, or they only apply to the case where skill levels are known. In our setting, we need to both estimate skills and assign workers to tasks given our uncertainty. Skill estimation is an interesting problem in itself. In this paper, we consider a feedback model that allows for a very simple scheme to be used. This allows us to separate the effects of the estimation from the matching algorithm itself.

\subsection{Skill estimation under threshold feedback}
\label{sec:multi_est_min_max}

Since we only obtain one rating for a set of skills, we introduce a method call \emph{min-max estimation}. 
Our estimate represents the minimum and maximum values that the algorithm believes the skill level could be. After each performed task, the algorithm uses the result to decide how to update the minimum and maximum estimations and then the estimation for that particular skill level is the average of the minimum and maximum values. It is easy to see that if the rewards are deterministic, then the following estimates will eventually become equal to the true worker's skills:
\begin{align}
  \label{eq:hie_min_update}
  s^{\min}_{\worker,i} &=
                \begin{cases}
                  \req_{\worker,i}, & r = 1 \\
                  s^{\min}_{\worker,i}, & \textrm{otherwise}
                \end{cases}
  \\
    \label{eq:hie_max_update}
    s^{\max}_{\worker,i}&=
                 \begin{cases}
                   \req_{\worker,i}, & r = 0 \\
                   s^{\max}_{\worker,i}, & \textrm{otherwise}
                 \end{cases},
\end{align}
where $s^{\min}_{\worker,i}$ and $s^{\max}_{\worker,i}$ are the minimum and maximum estimations for skill i respectively, $\req_{\task,i}$ is the $i$-th skill requirement for task $\task$, and $r$ is the rating received for the worker. The estimation for a particular skill is thus continually updated and set according to the following formula:
\begin{equation}
    \label{eq:hie_level_update}
    \widehat{\skill}_{\worker,i} = \dfrac{s^{\min}_{i} + s^{\max}_{i}}{2},
\end{equation}
where $s^{e}_{i}$ is the estimated skill level for skill $i$.

\begin{figure} 
    \centering
    \caption{Performance of the BEF solution, comparing the old skill estimation that uses the average of all ratings, and the new model that uses min-max estimation.}
    \label{fig:skill_est_old_vs_new}
    \begin{tikzpicture}
    \begin{axis}[
        width=0.45\textwidth,
        height=0.45\textwidth,
        title={},
        xlabel={Number of tasks},
        ylabel={Percent of optimal},
        xmin=10, xmax=300,
        ymin=64, ymax=86,
        xtick={10,25,50,75,100,125,150,175,200,225,250,275,300},
        ytick={64,66,68,70,72,74,76,78,80,82,84,86},
        ymajorgrids=true,
        grid style=dashed,
        legend pos=north west,
    ]
     
        \addplot coordinates {(10, 68.80753968253967)(20, 75.71051562816268)(30, 69.13597034818758)(40, 78.73163069547863)(50, 77.83444476061936)(60, 83.96402982538558)(70, 79.28204809648545)(80, 79.3220461290997)(90, 80.29548259104487)(100, 84.75000937362327)(110, 84.61906387138478)(120, 84.86243418503781)(130, 83.05979526836889)(140, 79.40791976091249)(150, 81.62697523614806)(160, 83.48911192337981)(170, 82.73717051670013)(180, 81.91732284035685)(190, 79.98643635185074)(200, 81.96784934539771)(210, 82.20838690676811)(220, 83.7916713744392)(230, 79.43889520205744)(240, 80.34551156142278)(250, 83.41464254171127)(260, 83.20300542903622)(270, 82.69982962747862)(280, 82.19177113624333)(290, 81.39886080475495)(300, 83.94144819074468)};
        \addlegendentry{min-max}
        \addplot coordinates {(10, 66.13690476190476)(20, 67.23203358389117)(30, 75.61644062785801)(40, 74.3191767461342)(50, 74.8662616846052)(60, 78.34538560691232)(70, 76.16035142726135)(80, 76.71646425056375)(90, 75.81605973949851)(100, 78.71874528804257)(110, 79.60901049565194)(120, 80.98534788163578)(130, 79.05473827909418)(140, 77.66604850265139)(150, 81.24913540807202)(160, 80.04697979593742)(170, 79.29066619296442)(180, 78.15321972753112)(190, 77.75843614882918)(200, 79.61460439373784)(210, 79.70343097603971)(220, 79.57329524951416)(230, 77.3757592460322)(240, 78.41685782261607)(250, 79.87135533893863)(260, 80.88129080770058)(270, 79.26198899829116)(280, 79.58134009042477)(290, 78.29611540105209)(300, 80.32899960444928)};
        \addlegendentry{average}
    \end{axis}
    
    \end{tikzpicture}
\end{figure}
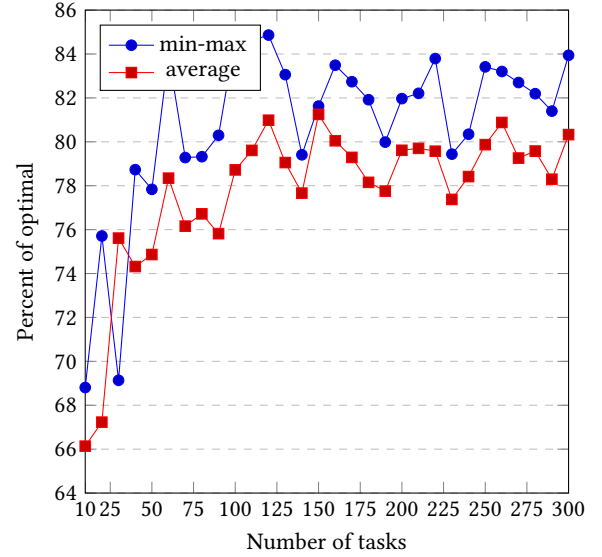

A comparison of the two rating methods can be seen in Figure \ref{fig:skill_est_old_vs_new}, and when comparing both to the optimal solution it is clear that using min-max estimates the true skill levels both better and faster.

\subsection{Hungarian min-max estimation}

The intuition behind Hungarian min-max estimation is that always making the optimal assignments with regards to all current workers and tasks will produce the best results, as long as the workers skill levels are continuously estimated. However, since the skills of workers are unknown, we simply plug in the min-max estimate.

\section{Experiments}
The experiments in this paper have been done on synthetic
data. Working purely with synthetic data requires some research about
the format of the relevant real world data, as well as a lot of
testing. To speed up computing times and to make analyzing the data
smoother, some restrictions were made:

\begin{itemize}
    \item The sets of workers skills and tasks requirements are always of equal size, which, we set to $3$. This is a minor restriction however, since both skill levels and requirement levels can be $0$, meaning the effective number of skills or requirements can be lower.
    \item The generated skill levels and requirement levels are sampled from a multinomial distribution with the categories $[0.0, 0.2, 0.4, 0.6, 0.8, 1.0]$, and all categories have the same probability to be chosen. Limiting the number of skill levels to $6$ was done to mimic real world data, since it would be an unrealistic expectation that humans would be able to estimate skill levels and requirement levels to any of an infinite number of levels between $0$ and $1$.
    \item The number of tasks available is equal to or greater than the number of available workers.
    \item All estimated skill levels are initiated to $0.5$, which is the average level when nothing is known.
\end{itemize}

\subsection{Algorithm paramaters}
\label{sec:res_alg_param}
When testing the different algorithms with the multi-skill matching model, a number of different parameters were used. Some of the parameters were the same for every test and algorithm but other were changed.

\begin{table}
\centering
\caption{Parameters used for testing all implemented methods.}
\label{Parameter-table}
\begin{tabular}{|l|l|}
\hline
\textbf{Parameter description} & \textbf{Value of parameter} \\ \hline
Number of workers & 10 \\ \hline
Number of tasks & 10 - 10000 \\ \hline
Number of runs & 25 \\ \hline
\end{tabular}
\end{table}

The number of workers was constant, and was set to $10$, as seen in table \ref{Parameter-table}, to speed up running time and to be able to analyze some of the results manually. Different number of tasks were used when testing different aspects. When trying to find the end result of assigning a large number of tasks, many tasks were generated as the running time was still withing acceptable limits. However, when trying to find how the performance changes with the number of tasks provided, many more tests had to be run and therefore fewer tasks were used. A run in this scenario is a single test with a set of tasks and a set of workers, and several runs were used because of two different elements of randomness with each run. First, when rating a worker's skill level, a Bernoulli distribution was used, and as described in Section~\ref{sec:multi_rating_model}, it has a certain probability of giving an erroneous rating. Second, since all worker skill levels and task requirement levels are generated uniformly at random, there is always a probability of two sets being generated where very few matches are possible, or none at all. To account for this, every test was run 25 times and the result presented is the average value of all those runs. Also, for each run, the set of tasks is constant while new workers are generated every time.

\subsection{Bernoulli rating}
\label{res:bern_rating}

All methods described in this paper use the same model for rating workers and their skill levels. As described before, this rating model includes a level of randomness in the form of $\varepsilon$, which affects the outcome. This is used to model several different aspects of uncertainty in a real life scenario:

\begin{itemize}
    \item The task provider's ability to accurately set the requirement level for each of the tasks requirements.
    \item The task provider's ability to accurately rate the performance of each skill level of a worker, compared to a task's requirement levels.
    \item The workers performance level. A worker can have a good or a bad day, resulting in a performance that is a poor representation of their true skill levels.
\end{itemize} 

\begin{figure}
    \centering
    \caption{Success rate of Hungarian min-max estimation with regard to $\varepsilon$, i.e. the probability that the correct skill level rating is given.}
    \label{fig:bern_eps}
    \begin{tikzpicture}
    \begin{axis}[
        width=0.45\textwidth,
        height=0.45\textwidth,
        title={},
        xlabel={$\varepsilon$},
        ylabel={Percent successrate},
        xmin=0, xmax=1,
        ymin=30, ymax=80,
        xtick={0.0,0.1,0.2,0.3,0.4,0.5,0.6,0.7,0.8,0.9,1.0},
        ytick={30,40,50,60,70,80},
        ymajorgrids=true,
        grid style=dashed,
        legend pos=south east,
    ]
     
        \addplot coordinates {(0.0, 70.69333333333334)(0.1, 62.58666666666666)(0.2, 57.37333333333332)(0.3, 54.426666666666655)(0.4, 51.946666666666665)(0.5, 49.45333333333334)(0.6, 46.45333333333333)(0.7, 46.78666666666666)(0.8, 43.62666666666667)(0.9, 42.68)(1.0, 39.733333333333334)};
    \end{axis}
    
    \end{tikzpicture}
\end{figure}
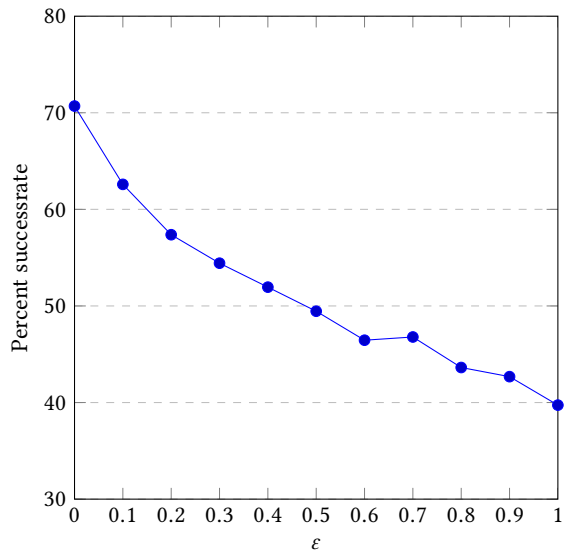

A number of different values of $\varepsilon$ were tested, using Hungarian min-max estimation, to see how the overall result was affected. The results of the test can be seen in Figure \ref{fig:bern_eps}.

The value for $\varepsilon$ was set to $0.15$ to represent all the different sources of error described above. In this case it means that there is a $15\%$ probability that an error occurs during the process of task creation, skill rating och worker performance.

\subsection{Performance compared to optimal}
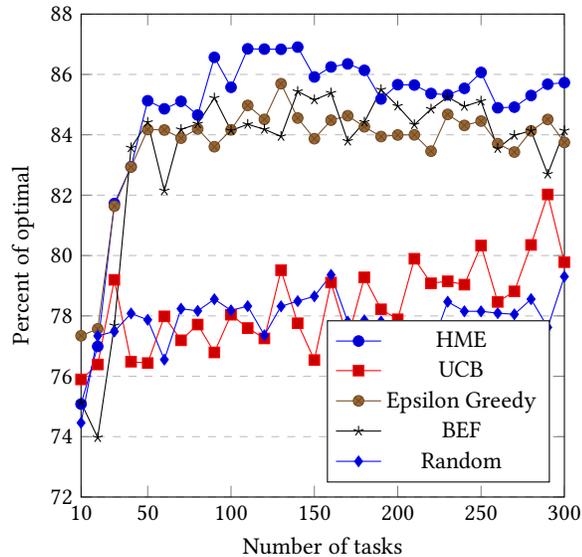
\begin{figure}
    \centering
    \caption{Performance of all algorithms compared to the optimal solution. Used to see if any of them converge towards the optimal solution, given enough tasks. Maximum number of skills per worker and requirements per task is $3$.}
    \label{fig:regression_all}
    \begin{tikzpicture}
    \begin{axis}[
        width=0.45\textwidth,
        height=0.45\textwidth,
        title={},
        xlabel={Number of tasks},
        ylabel={Percent of optimal},
        xmin=10, xmax=300,
        ymin=72, ymax=88,
        xtick={10,50,100,150,200,250,300},
        ytick={72,74,76,78,80,82,84,86,88},
        ymajorgrids=true,
        grid style=dashed,
        legend pos=south east,
    ]
     
        \addplot coordinates {(10, 75.07842055732216)(20, 76.98833558430668)(30, 81.72335937688335)(40, 82.94135887732747)(50, 85.12815416024979)(60, 84.8598376030712)(70, 85.10809386264096)(80, 84.65456310892765)(90, 86.56665767151154)(100, 85.57297059769567)(110, 86.84799966913361)(120, 86.83912010234418)(130, 86.83267877247582)(140, 86.90502981765636)(150, 85.91621508416405)(160, 86.2474915652567)(170, 86.34831593818474)(180, 86.13588077835763)(190, 85.18576604470286)(200, 85.66097867769392)(210, 85.64797324364088)(220, 85.368336165826)(230, 85.32307752050423)(240, 85.53801911989953)(250, 86.06613321881152)(260, 84.89407756411558)(270, 84.9155819999599)(280, 85.3016888419201)(290, 85.67510894323857)(300, 85.7284892164905)};
        \addlegendentry{HME}
        \addplot coordinates {(10, 75.89583180612931)(20, 76.39029331819766)(30, 79.19453333337542)(40, 76.48142538847685)(50, 76.43798524135283)(60, 77.98533427778)(70, 77.19234925429086)(80, 77.71364828786565)(90, 76.78907027722175)(100, 78.03883112054775)(110, 77.5987564572628)(120, 77.25406155793456)(130, 79.51096411270271)(140, 77.75694573623441)(150, 76.53755680371329)(160, 79.10495135034785)(170, 77.24281618858797)(180, 79.27864145009478)(190, 78.21920175425497)(200, 77.88838271941833)(210, 79.89281439626441)(220, 79.08140720984163)(230, 79.14731381810623)(240, 79.03550984284365)(250, 80.3321672129217)(260, 78.46302476165114)(270, 78.81327304445686)(280, 80.35138014671708)(290, 82.02372910904111)(300, 79.77896643927687)};
        \addlegendentry{UCB}
        \addplot coordinates {(10, 77.34089750066866)(20, 77.5691976767938)(30, 81.63742351515872)(40, 82.93350783385814)(50, 84.17517204903156)(60, 84.1591218686784)(70, 83.88916654708021)(80, 84.19777854738048)(90, 83.60494936616057)(100, 84.16393620708688)(110, 84.97294138946332)(120, 84.51074273909744)(130, 85.69259889185203)(140, 84.55485711530417)(150, 83.87079146886987)(160, 84.48366361118154)(170, 84.63081889508062)(180, 84.26500234596371)(190, 83.94169338982536)(200, 83.99376865578438)(210, 83.99853827514119)(220, 83.45654339312118)(230, 84.67636508031332)(240, 84.31063141585864)(250, 84.4540586896958)(260, 83.70624542985172)(270, 83.42825100339452)(280, 84.12856879337009)(290, 84.50824507009644)(300, 83.74750393262376)};
        \addlegendentry{Epsilon Greedy}
        \addplot coordinates {(10, 75.13017367199289)(20, 73.96811427607476)(30, 77.68206728707135)(40, 83.56980695994129)(50, 84.41251978385331)(60, 82.14854770425893)(70, 84.17677052218212)(80, 84.36260942400786)(90, 85.22021627104454)(100, 84.13944744916371)(110, 84.34739012384114)(120, 84.18192851837895)(130, 83.9471500877767)(140, 85.43631869334473)(150, 85.15866282690912)(160, 85.38588482634862)(170, 83.79077679933916)(180, 84.42068567357617)(190, 85.493933540555)(200, 84.95129455012254)(210, 84.33637515084502)(220, 84.85181408268858)(230, 85.25188725240241)(240, 84.94332924950263)(250, 85.1172598725988)(260, 83.54776641186686)(270, 83.9821103669956)(280, 84.13265038597599)(290, 82.69832675221846)(300, 84.13723192088568)};
        \addlegendentry{BEF}
        \addplot coordinates {(10, 74.45840338659562)(20, 77.33819769747812)(30, 77.4836661929069)(40, 78.08195967047024)(50, 77.87007389729496)(60, 76.5501830794159)(70, 78.23765229478359)(80, 78.16622905967701)(90, 78.55239271833841)(100, 78.18333446994923)(110, 78.32274361244315)(120, 77.359079941694)(130, 78.31776278300582)(140, 78.48925957113049)(150, 78.6489839117215)(160, 79.36013290530546)(170, 77.81059833513642)(180, 77.85540711620655)(190, 77.81691842236546)(200, 77.41524735546393)(210, 77.43967994584925)(220, 76.9513914585328)(230, 78.4689190756209)(240, 78.15425375691976)(250, 78.15166424480687)(260, 78.091640615201)(270, 78.05230137299608)(280, 78.55576639761235)(290, 77.62212866798124)(300, 79.3036300631313)};
        \addlegendentry{Random}
    \end{axis}
    
    \end{tikzpicture}
\end{figure}

When all methods were implemented, using block matching, they were compared to the optimal solution over time, as an increasing number of tasks were assigned. The goal for any method is to make sure that the estimated skill levels of every worker converges towards the true skill levels, and each method uses the same set of workers and tasks. During most of the experimentation, the number of available skills and requirements were limited to $3$, to speed up computing time, while still using more than $1$ skill. The results of all methods, while using $3$ skills and requirements, can be seen in Figure \ref{fig:regression_all}.

Tests made with more skills and requirements, from $3$ up to $10$, all give similar results as with 3 skills, meaning the big difference occurs when changing from $1$ skill and requirement to multiple.

\section{Discussion}
\label{sec:discussion}
In this paper, we explored some initial algorithms towards solving the generalised matching problem under uncertainty. While the matching problem has been studied extensively under perfect information, a lot remains to be done for the case of imperfect information. While some special cases of this problem can be thought of as bandit problems, in general it exhibits a much larger complexity, so standard bandit algorithms are not directly applicable. In particular, even if we have multiple skills in each task, feedback received is only corresponding to overall worker performance. 

Our solution entailed the assumption of a specific feedback model, and very simple point estimates of worker skills, combined with the well-known Hungarian algorithm.  However, we believe that much more sophisticated algorithms and models could be brought to bear upon this important problem. On the algorithmic side, we could use proper confidence bounds (similar to those used in contextual bandit problems). On the modelling side, it would be interesting to consider social networking between workers, their relationship between different skills, as well as the reputation of employers, in a general graphical model setting. 

Another open question is what the appropriate feedback model is. In our work, we  considered a model where the probability of positive feedback only depended on whether a worker satisfied a set of skills. 
Nevertheless, we believe that this preliminary research sets the scene for plenty of follow-up work. We have shown that even very simple algorithms can comfortably beat naive methods that do not take into account the problem structure. We believe it should be possible to derive analogues of most well-known bandit algorithms for this new problem, and provide appropriate performance guarantees, something that would be highly beneficial to workers and employers alike.

\paragraph{Acknowledgements.}
The research has received funding from: the People Programme (Marie Curie Actions) of the European Union's Seventh Framework Programme (FP7/2007-2013) under REA grant agreement 608743, the Future of Life Institute, and the Swiss National Science Foundation.

\bibliographystyle{plainnat}
\bibliography{references}

\end{document}